\def\year{2020}\relax
\documentclass[letterpaper]{article} 
\usepackage{aaai20}  
\usepackage{times}  
\usepackage{helvet} 
\usepackage{courier}  
\usepackage[hyphens]{url}  
\usepackage{graphicx} 
\usepackage{graphicx}  
\usepackage{amsmath} 
\usepackage{amssymb} 
\usepackage{multirow} 
\usepackage{pifont} 
\newcommand{\xmark}{\ding{55}} 
\newcommand{\cmark}{\ding{51}} 
\usepackage{csquotes} 
\title{FISR: Deep Joint Frame Interpolation and \\Super-Resolution with a Multi-scale Temporal Loss}
\author{\Large \textbf{Soo Ye Kim,\footnotemark[1]} \Large \textbf{Jihyong Oh,}\thanks{Both authors contributed equally to this work.} \Large \textbf{Munchurl Kim}\\ 
Korea Advanced Institute of Science and Technology\\
Republic of Korea\\
\{sooyekim, jhoh94, mkimee\}@kaist.ac.kr
}

\begin{document}

\maketitle

\begin{abstract}
Super-resolution (SR) has been widely used to convert low-resolution legacy videos to high-resolution (HR) ones, to suit the increasing resolution of displays (e.g. UHD TVs). However, it becomes easier for humans to notice motion artifacts (e.g. motion judder) in HR videos being rendered on larger-sized display devices. Thus, broadcasting standards support higher frame rates for UHD (Ultra High Definition) videos (4K@60 fps, 8K@120 fps), meaning that applying SR only is insufficient to produce genuine high quality videos. Hence, to up-convert legacy videos for realistic applications, not only SR but also video frame interpolation (VFI) is necessitated. In this paper, we first propose a joint VFI-SR framework for up-scaling the spatio-temporal resolution of videos from 2K 30 fps to 4K 60 fps. For this, we propose a novel training scheme with a multi-scale temporal loss that imposes temporal regularization on the input video sequence, which can be applied to any general video-related task. The proposed structure is analyzed in depth with extensive experiments. 
\end{abstract}

\section{Introduction}
With the prevalence of high resolution (HR) displays such as UHD TVs or 4K monitors, the demand for higher resolution visual contents (videos) is also increasing with $\text{YouTube}^{\text{TM}}$ already supporting 8K UHD video services (7680$\times$4320). Super-resolution (SR) technologies are closely related to this trend, as they can enlarge the spatial resolution of legacy low resolution (LR) videos to higher resolution ones. However, the increase in spatial resolution necessarily entails the increase in \textit{temporal} resolution, or the \textit{frame rate}, for videos to be properly rendered on larger-sized displays from a perceptual quality perspective. The human visual system (HVS) becomes more sensitive to the temporal distortion of videos as the spatial resolution increases, and tends to easily perceive motion judder (discontinuous motion) artifacts in HR videos, which deteriorates the perceptual quality \cite{daly2001engineering}. To this regard, the frame rate must be increased from low frame rate (LFR) to high frame rate (HFR) for HR videos to be visually pleasing. This is the reason behind UHD (Ultra High Definition) broadcast standards specifying 60 fps and 120 fps (frames per second) for 4K (3840$\times$2160) and 8K (7680$\times$4320) UHD videos \cite{dvb-uhd}, compared to the 30 fps of conventional 2K (FHD, 1920$\times$1080) videos.

\begin{figure} [t]
\centering
\includegraphics[width=\columnwidth]{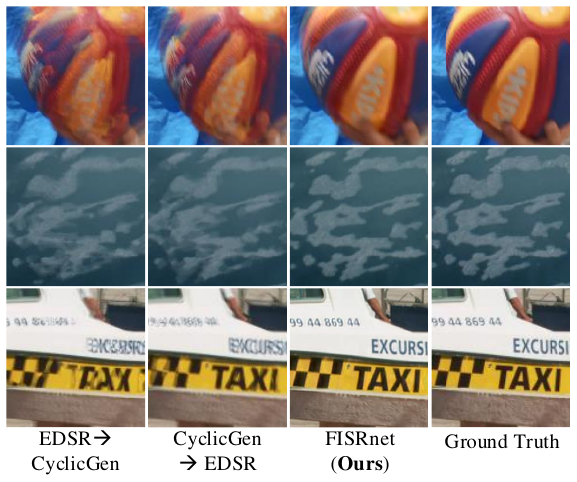}
\caption{Qualitative comparison with the cascade of existing methods. The proposed FISRnet is able to reconstruct the texture of moving waters and small letters on objects.}
\label{fig:1}
\end{figure}

Therefore, in order to convert legacy 2K 30 fps videos to genuine 4K 60 fps videos that can be viewed on 4K UHD displays, video frame interpolation (VFI) is essential along with SR. Nevertheless, VFI and SR have been intensively but separately studied in low level vision tasks. None of the existing methods have jointly handled both VFI and SR problems, which is a complex task where both the spatial and temporal resolutions must be increased. In this paper, we first propose a joint VFI-SR method, called FISR, that enables the direct conversion of 2K 30 fps videos to 4K 60 fps. We employ a novel training strategy that handles multiple consecutive samples of video frames per each iteration, with a novel temporal loss that exerts temporal regularization across these consecutive samples. This scheme is general and can be applied to any video-related task. To handle the high resolution of 4K UHD, we propose a multi-scale structure trained with the novel temporal loss applied across all scale levels.

\smallskip
Our contribution can be summarized as follows:
\begin{itemize}
\item We first propose a joint VFI-SR method that can simultaneously increase the spatio-temporal resolution of video sequences.
\item We propose a novel multi-scale temporal loss that can effectively regularize the spatio-temporal resolution enhancement of video frames with high prediction accuracy.
\item All our experiments are based on 4K 60 fps video data to account for realistic application scenarios.
\end{itemize}

\section{Related Work} 
\subsection{Video Super-Resolution}
\begin{figure} [t]
\centering
\includegraphics[scale=0.62]{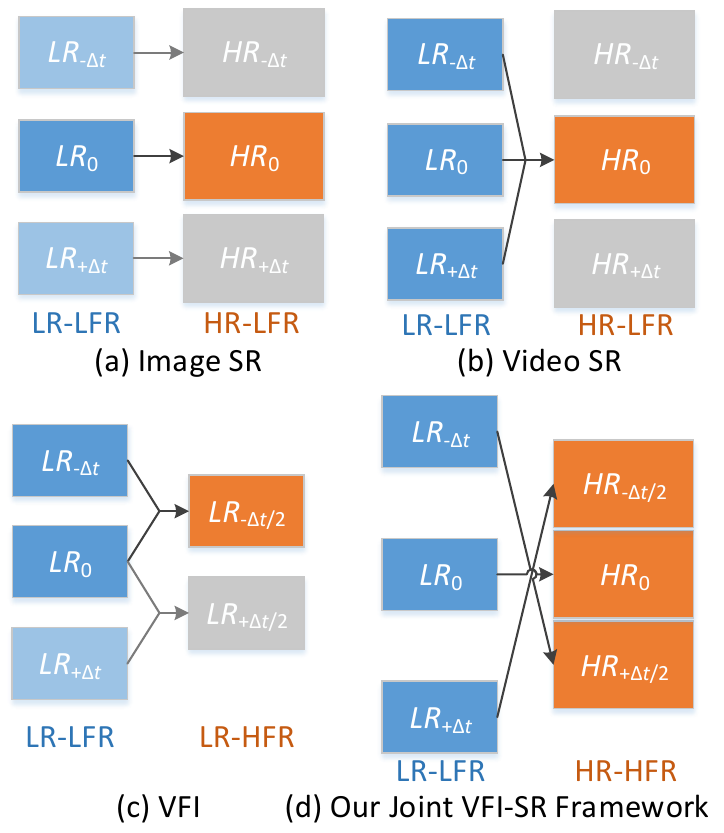}
\caption{Comparison between common VFI and SR frameworks and our joint VFI-SR framework.}
\label{fig:2}
\end{figure}

The purpose of SR is to recover the lost details of the LR image to reconstruct its HR version. SR is widely used in diverse areas such as medical imaging \cite{yang2012coupled}, satellite imaging \cite{cao2016towards}, and as pre-processing in person re-identification \cite{jiao2018deep}. With the recent success of deep-learning-based methods in computer vision, various single-image SR (SISR) methods have been developed \cite{dong2015image,lim2017enhanced,LapSRN,zhang2019image}, which enhance the spatial resolution by focusing only on the spatial information of the given LR image as shown in Fig. \ref{fig:2} (a).

On the other hand, video SR (VSR) can additionally utilize the temporal information of the consecutive LR frames to enhance the performance. If SISR is independently applied to each of the single frames to generate the VSR results, the output HR videos tend to lack temporal consistency, which may cause flickering artifacts \cite{shi2016real}. Therefore, VSR methods exploit the additional temporal relationships as in Fig. \ref{fig:2} (b), and popular ways to achieve this include simply concatenating the sequential input frames, or adopting 3D convolution filters \cite{caballero2017real,huang2017video,jo2018deep,li2019fast,kim2019}. However, these methods tend to fail to capture large motion, where the absolute motion displacements are large, or multiple local motions, due to the simple concatenation of inputs where many frames are processed simultaneously in the earlier part of the network. Furthermore, the use of 3D convolution filters leads to expensive computation complexity, which may cause the degradation of VSR performance when the overall network capacity is restricted. 
To overcome this issue, various methods have utilized motion information \cite{makansi2017end,wang2018learning,Kalarot_2019_CVPR_Workshops}, especially optical flow, to improve the prediction accuracy. While using motion information, Haris \textit{et al}. \cite{haris2019recurrent} proposed an iterative refinement framework to combine the spatio-temporal information of LR frames by using a recurrent encoder-decoder module. It is worth pointing out that although Vimeo-90K \cite{xue2017video} with $448\times256$ resolution is a relatively high resolution benchmark dataset used in VSR, it is still insufficient to represent the characteristics of recent UHD video data. Furthermore, none of the aforementioned VSR methods generate HFR frames simultaneously.

\subsection{Video Frame Interpolation}
The goal of VFI is to generate high quality non-existent middle frames by appropriately combining two original consecutive input frames as in Fig. \ref{fig:2} (c). VFI is highly important in video processing as viewers tend to feel visually comfortable towards HFR videos \cite{mackin2015}. VFI has been applied to various applications such as slow motion generation \cite{jiang2018super}, frame rate up conversion (FRUC) \cite{yu2019hierarchical}, novel view synthesis \cite{flynn2016deepstereo}, and frame recovery in video streaming \cite{wu2015modeling}. The main difficulties in VFI are the consideration of fast object motion and the occlusion problem. Fortunately, with various deep-learning-based methods, VFI has been actively studied and has shown impressive results on LR benchmarks  \cite{niklaus2018context,liu2019deep,bao2019depth}. Niklaus \textit{et al}. \cite{niklaus2018context} proposed a context-aware frame synthesis method where per-pixel context maps are extracted and warped prior to entering a GridNet architecture for enhanced frame interpolation. Liu \textit{et al}. \cite{liu2019deep} proposed a cycle consistency loss that not only forces the network to enhance the interpolation performance, but also makes better use of the training data. Bao \textit{et al}. \cite{bao2019depth} proposed DAIN, which jointly optimizes five different network components to produce a high quality intermediate frame by exploiting depth information. 

However, these methods face difficulties against higher resolution videos, where the absolute motion tends to be large, often exceeding the receptive field of the networks, resulting in performance degradation of the interpolated frames. Meyer \textit{et al}. \cite{Meyer2015} first noticed the weakness of VFI methods for HR videos, and employed a hand-crafted phase-based method. Among deep-learning-based methods, a deep CNN was proposed in IM-Net \cite{peleg2019net} to cover fast motions so that it can handle the VFI for higher resolution ($1344\times768$) inputs. However, their testing scenarios were still limited to spatial resolutions lower than 2K videos, which is not adequate for 4K/8K TV displays.

On the other hand, Ahn \textit{et al}. \cite{ahn2019fast} first proposed a hybrid task-based network for a fast and accurate VFI of 4K videos based on a coarse-to-fine approach. To reduce the computation complexity, they first down-sample two HR input frames prior to temporal interpolation (TI), and generate an LR version of the interpolated frame. Then, a spatial interpolation (SI) takes in the bicubic up-sampled version of the LR interpolated frame concatenated with the original two HR input frames to synthesize the final VFI output. Although their network performs a two-step spatio-temporal resolution enhancement, it should be noted that they take an advantage of the original 4K input frames, and their final goal is VFI (not joint VFI-SR) of 4K videos. This is different from our problem of jointly optimizing VFI-SR that generates the HR-HFR outputs directly from the LR-LFR inputs.

In this paper, we handle the joint VFI-SR, especially for FRUC applications, to generate high quality middle frames with higher spatial resolutions, which enables the direct conversion of 2K 30 fps videos to 4K 60 fps videos, named as frame interpolation and super-resolution (FISR). This is a novel problem, which has not been previously considered.

\section{Proposed Method}
\subsection{Input/Output Framework}
A common VFI framework involves the prediction of a single middle frame from the input of two consecutive frames as in Fig. \ref{fig:2} (c). In this case, the final HFR video constitutes of alternately located original input frames between the interpolated frames. However, this scheme cannot be directly applied for joint VFI-SR since the spatial resolutions of original input frames (LR) and predicted frames (HR) are different, and there is a resolution mismatch if we wish to insert the input frames among the predicted frames. Therefore, we propose a novel input/output framework as shown in Fig. \ref{fig:2} (d), where three consecutive HR HFR frames are predicted from three consecutive LR LFR frames. That is, for every three consecutive LR input frames, only SR is performed to produce the middle HR$_{0}$ output frame while joint VFI-SR is performed to synthesize the other two end-frames (HR$_{-\Delta t/2}$ and HR$_{+\Delta t/2}$). With the per-frame shift of a sliding temporal window, the frames HR$_{-\Delta t/2}$ and HR$_{+\Delta t/2}$ in the current temporal window will overlap with HR$_{+\Delta t/2}$ from the previous temporal window, and HR$_{-\Delta t/2}$ from the next time window, respectively. As blurry frames are produced if the two overlapping frames are averaged, the frame from the later sliding window is used for simplicity.

\subsection{Temporal Loss}
\begin{figure*}
\centering
\includegraphics[width=\textwidth]{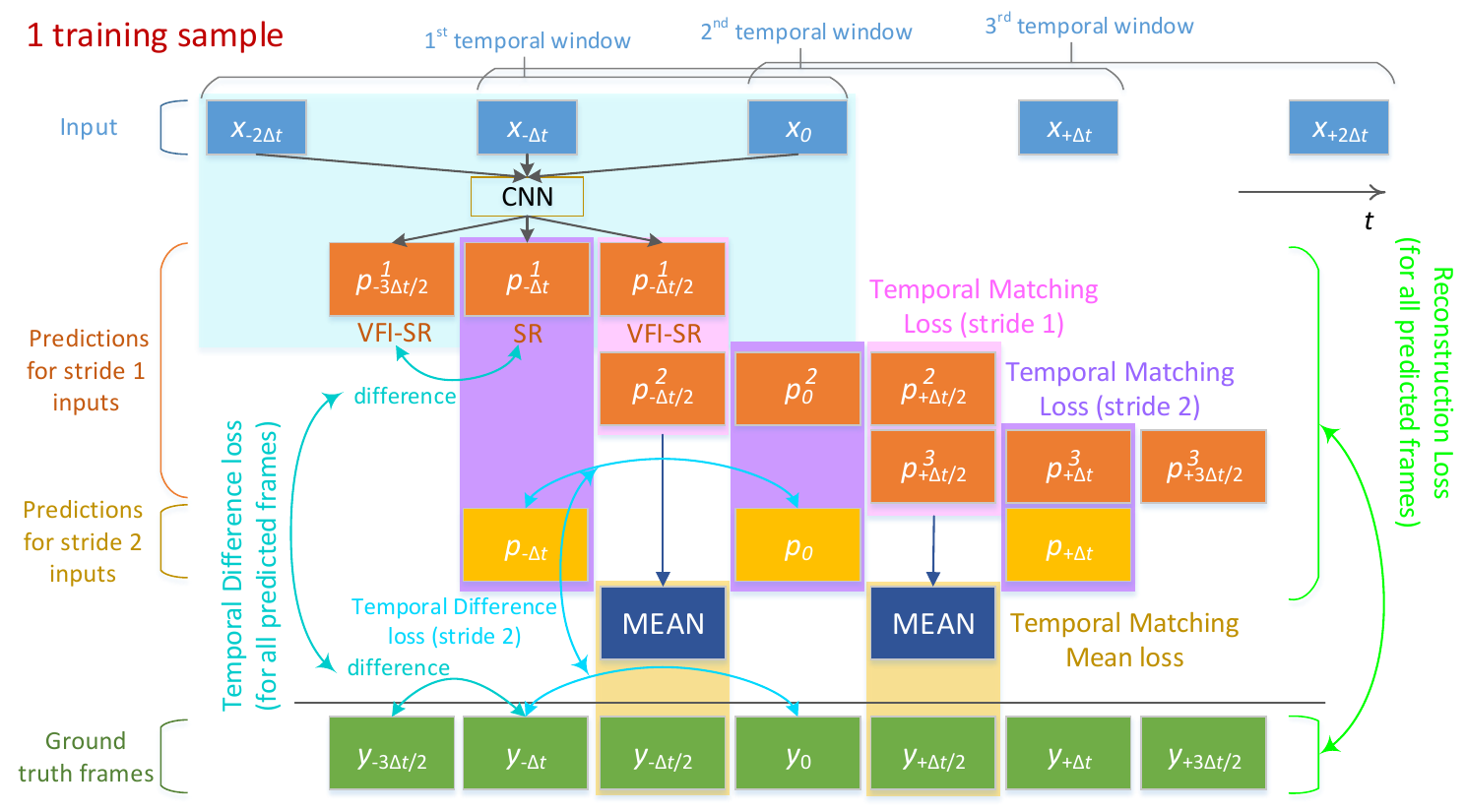}
\caption{Temporal Loss}
\label{fig:3}
\end{figure*}

We propose a novel temporal loss for regularization in network training with video sequences. Instead of back-propagating the error at each mini-batch of data samples of three input/predicted frames, a training sample of FISR is composed of five consecutive input frames, thus containing three consecutive data samples with temporal stride 1 and one data sample with temporal stride 2. By considering the relations of these multiple data samples, more regularization is \textit{temporally} imposed on network training for a more stable prediction. A detailed schema of this \textit{multiple data sample} training strategy is illustrated in Fig. \ref{fig:3}.

As shown in Fig. \ref{fig:3}, we let the input frames be $x_{t}$, where \textit{t} is the time instance. Then, one training sample consists of five frames, $\{x_{-2\Delta t}, x_{-\Delta t}, x_{0},$ $x_{+\Delta t}, x_{+2\Delta t}\}$, and each training sample includes \textit{three} data samples with temporal stride 1, $\{x_{-2\Delta t}, x_{-\Delta t}, x_{0}\},$ $\{x_{-\Delta t}, x_{0}, x_{+\Delta t}\}, \{x_{0}, x_{+\Delta t}, x_{+2\Delta t}\}$ at each temporal window centered at $-\Delta t$, 0, and $+\Delta t$, respectively. Their corresponding predictions are denoted by $p_{t}^{w}$, where $w$ indicates the $w$-th temporal window, and their ground truth frames are given by $y_{t}$.

\subsubsection{Temporal Matching Loss}
Due to the sliding temporal window within each training sample, there exist two time instances $-\Delta t/2$ and $+\Delta t/2$ where the predicted frames overlap across the different time window \textit{w}. The temporal matching loss enforces these overlapping frames to be similar to each other, formally given by,
\begin{multline} 
	L_{TM}^{1} = \lVert p_{-\Delta t/2}^{1}-p_{-\Delta t/2}^{2} \rVert_{2}+\lVert p_{+\Delta t/2}^{2}-p_{+\Delta t/2}^{3}\rVert_{2}.
\end{multline}

We also consider an additional data sample with temporal stride 2 within the training sample, $\{x_{-2\Delta t}, x_{0}, x_{+2\Delta t}\}$ centered at 0,  which in turn produces $\{p_{-\Delta t}$, $p_{0}$, $p_{+\Delta t}\}$, as shown in \textit{yellow boxes} in Fig. \ref{fig:3}. With the stride 2 predictions, there are three overlapping time instances with the predictions from the stride 1 data samples. Accordingly, the temporal matching loss for stride 2 is given by,
\begin{multline} 
	L_{TM}^{2} = \lVert p_{-\Delta t}-p_{-\Delta t}^{1} \rVert_{2}+\lVert p_{0}-p_{0}^{2}\rVert_{2}+\lVert p_{+\Delta t}-p_{+\Delta t}^{3} \rVert_{2}.
\end{multline}

\subsubsection{Temporal Matching Mean Loss}
To further regularize the predictions, we also impose the L2 loss on the mean of the overlapping frames of stride 1 and the corresponding ground truth at the overlapping time instance as follows: 
\begin{multline} 
	L_{TMM} = \lVert\frac{1}{2}(p_{-\Delta t/2}^{1}+p_{-\Delta t/2}^{2})-y_{-\Delta t/2}\rVert_{2}+ \\ \lVert\frac{1}{2}(p_{+\Delta t/2}^{2}+p_{+\Delta t/2}^{3})-y_{+\Delta t/2}\rVert_{2}.
\end{multline}

\begin{figure} [t]
\centering
\includegraphics[width=\columnwidth]{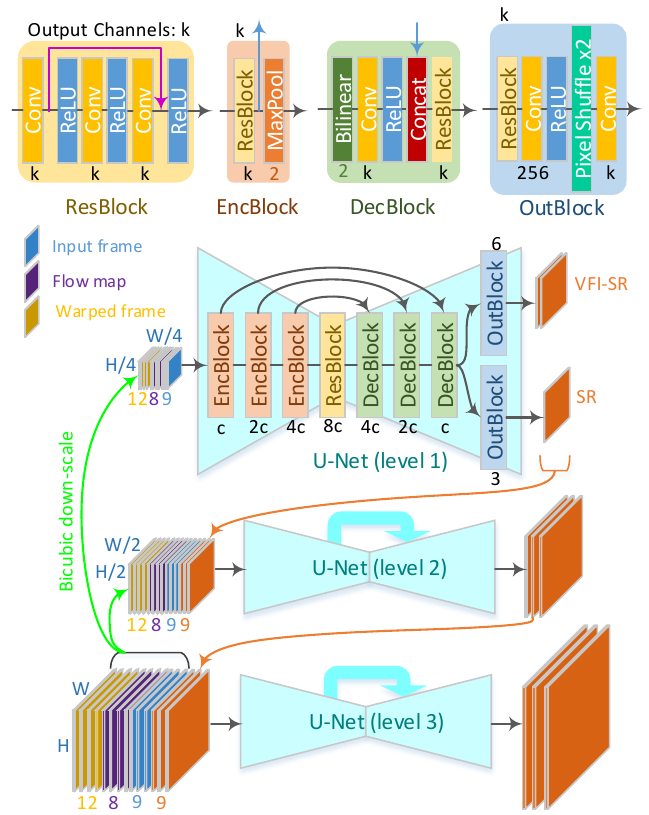}
\caption{Network Architecture}
\label{fig:4}
\end{figure}

\subsubsection{Temporal Difference Loss}
In order to enforce the temporal coherence in the predicted frames, we design a simple temporal difference loss, $L_{TD}$, applied for all sets of predictions, where the difference between the consecutive predicted frames must be similar to the difference between the consecutive ground truth frames. For the predictions from the data samples of temporal stride 1, the loss is given by,
\begin{multline}
	L_{TD}^{1} = \sum_{w=1}^{3}\sum_{s=0}^{1}\lVert (p_{(w+\frac{s}{2}-\frac{5}{2})\Delta t}^{w}-p_{(w+\frac{s}{2}-2)\Delta t}^{w})-\\ (y_{(w+\frac{s}{2}-\frac{5}{2})\Delta t}-y_{(w+\frac{s}{2}-2)\Delta t})\rVert_{2}.
\end{multline}
For the stride 2 predictions, the loss is given by,
\begin{multline}
	L_{TD}^{2} = \lVert (p_{-\Delta t}-p_{0})-(y_{-\Delta t}-y_{0})\rVert_{2}+\\ \lVert (p_{0}-p_{+\Delta t})-(y_{0}-y_{+\Delta t})\rVert_{2}.
\end{multline}

\subsubsection{Reconstruction Loss}
Lastly, the reconstruction loss, $L_{R}$, is the L2 loss between all predicted frames and the corresponding ground truths. Firstly, for the predictions from the data samples of temporal stride 1, the loss is given as,
\begin{align}
	L_{R}^{1} = \sum_{w=1}^{3}\sum_{s=0}^{2}\lVert p_{(w+\frac{s}{2}-\frac{5}{2})\Delta t}^{w}-y_{(w+\frac{s}{2}-\frac{5}{2})\Delta t}\rVert_{2}.
\end{align}
For the stride 2 predictions, the loss is given as,
\begin{align}
	L_{R}^{2} = \sum_{s=0}^{2}\lVert p_{(s-1)\Delta t}-y_{(s-1)\Delta t}\rVert_{2}.
\end{align}

\subsubsection{Total Loss}
Finally the total loss $L_{T}$ is given by,
\begin{multline}
	L_{T} = \lambda_{R}\cdot L_{R}^{1}+\lambda_{TM}^{1}\cdot L_{TM}^{1}+\lambda_{TMM}\cdot L_{TMM}+\lambda_{TD}\cdot L_{TD}^{1}\\+\lambda^{2}\cdot(\lambda_{R}\cdot L_{R}^{2}+\lambda_{TM}^{2}\cdot L_{TM}^{2}+\lambda_{TD}\cdot L_{TD}^{2}),
\label{eq8}
\end{multline}
where the different types of $\lambda$ are the weighting parameters for the corresponding losses to be determined empirically. The CNN parameters are updated at once for every mini-batch of \textit{training samples}, consisting of four data samples (three stride 1 samples and one stride 2 sample).

\subsection{Network Architecture}
We design a 3-level multi-scale network as shown in Fig. \ref{fig:4}, which is beneficial in handling large motion in the HR frames with the enlarged effective receptive fields in the lower scale levels. In levels 1 and 2, the input frames are down-scaled by 4 and 2, respectively, from level 3 of the original scale using a bicubic filter, and all scales employ the same U-Net-based architecture. With the multi-scale structure, a coarse prediction is generated at the lowest scale level, which is then concatenated and progressively refined at the subsequent scale levels. The total loss of Eq. \eqref{eq8} is respectively computed at all three scale levels $l\in\{1, 2, 3\}$ with weighting parameters $\lambda_{l}$, as $L=\sum_{l=1}^{3}\lambda_{l}\cdot L_{T}^{l}$. 

Furthermore, to effectively handle large motion and occlusions, the bidirectional optical flow maps and the corresponding warped frames are stacked with the input frames. We use the pre-trained PWC-Net \cite{Sun_2018_CVPR} to obtain the optical flows $\{f_{-\Delta t\rightarrow 0}$, $f_{0\rightarrow -\Delta t}$, $f_{0\rightarrow +\Delta t}$, $f_{+\Delta t\rightarrow 0}\}$, and the concatenated flow maps $\{f_{-\Delta t/2\rightarrow 0}$, $f_{-\Delta t/2\rightarrow -\Delta t}$, $f_{+\Delta t/2\rightarrow +\Delta t}$, $f_{+\Delta t/2 \rightarrow 0}\}$ are approximated with the linear motion assumption from the respective flow maps (e.g. $f_{-\Delta t/2\rightarrow 0} = 1/2\cdot f_{-\Delta t\rightarrow 0}$). The corresponding backward warped frames $\{g_{-\Delta t/2\rightarrow 0}$, $g_{-\Delta t/2\rightarrow -\Delta t}$, $g_{+\Delta t/2\rightarrow +\Delta t}$, $g_{+\Delta t/2 \rightarrow 0}\}$ are estimated from the approximated flows, and are also concatenated along with the input frames.

\section{Experiment Results}

\subsection{Experiment Conditions}
\subsubsection{Implementation Details}
All convolution filters have a kernel size of $3\times3$, and in the U-Net architecture, the output channel $c$ is set to 64. The final output channels are 6 for the two VFI-SR frames and 3 for the single SR frame. As PWC-Net was trained for RGB frames, the flows $f_{t}$ and the warped frames $g_{t}$ were obtained in the RGB domain, and the warped frames were converted back to YUV for concatenation with the input frames. For all experiments, the scale factor is 2 for both the spatial resolution and the frame rate, to target 2K 30 fps to 4K 60 fps applications, and we use Tensorflow 1.13 in our implementations.
\subsubsection{Data}
We collected 4K 60 fps videos of total 21,288 frames that contain 112 scenes with diverse object and camera motions from $\text{YouTube}^{\text{TM}}$. Among the collected scenes, we especially selected 88 for training and 10 scenes for testing, both of which contain large object or camera motions. In the 10 scenes for testing, the pixel displacement range amounts up to [-124, 109] in pixels, frame-to-frame, and all 10 scenes contain at least [-103, 94] pixel displacement within the input 2K video frame, quantitatively demonstrating the large motion contained in the data. Additionally, the average motion magnitude in each scene of the 2K frames ranges from 5.61 to 11.40 pixels frame-to-frame, with the total average for all 10 scenes being 7.64 pixels.

To create one training sample, we randomly cropped a series of $192\times192$ HR patches at the same location throughout 9 consecutive frames. With the 5-frame input setting as shown in Fig. \ref{fig:3}, the 2nd ($-3\Delta t/2$) to the 8th ($+3\Delta t/2$) frames were used as the 4K ground truth HR HFR frames, and the five odd-positioned frames ($-2\Delta t$, $-\Delta t$, $0$, $+\Delta t$, $+2\Delta t$) were bicubic down-scaled to the size of $96\times96$ to be used as the LR LFR input frames for training, as shown in \textit{green} and \textit{blue} boxes, respectively, in Fig. \ref{fig:3}. To obtain diverse training samples, each training sample was extracted with a frame stride of 10. By doing so, we constructed 10,086 training samples in total before starting the training process to avoid heavy training time required for loading 4K frames at every iteration. 

During the test phase, the test set composed of 10 different scenes with 5 consecutive LR (2K) LFR (30 fps) frames was used, where the full 2K frames were entered as a whole, and the average PSNR and SSIM were measured for a total of 90 ($=3$(two VFI-SR and one SR frame)$\times3$(three sliding windows in five consecutive input frames)$\times10$(ten scenes)) predicted frames. The input and the ground truth frames are in YUV channels, and the performance was also measured in the YUV color space.

\begin{figure}
\centering
\includegraphics[width=\columnwidth]{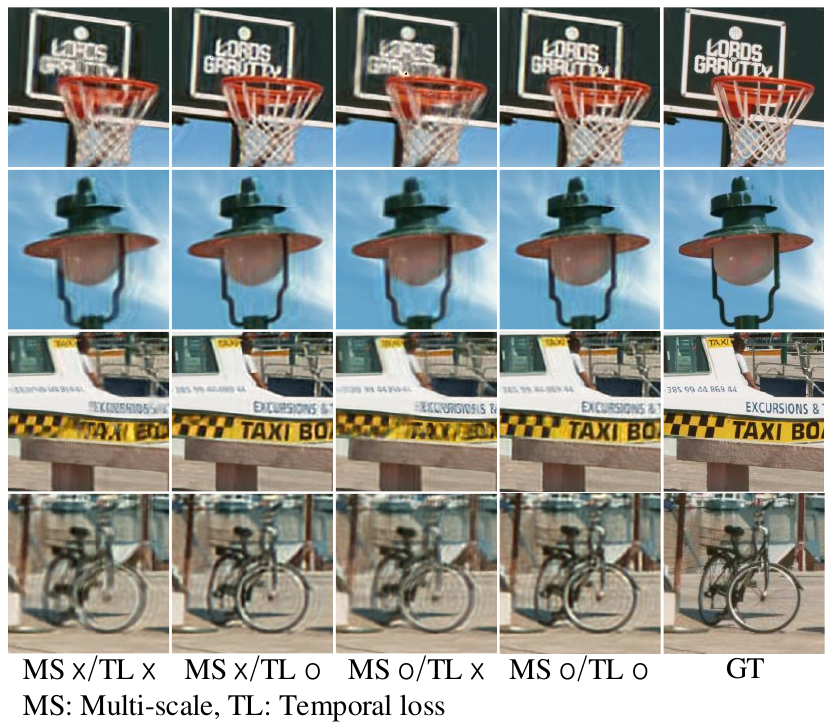}
\caption{Effect of the Temporal Loss}
\label{fig:5}
\end{figure}

\subsubsection{Training}
For training, we adopted the Adam optimizer \cite{kingma2014adam} with the initial learning rate of $10^{-4}$, reduced by a factor of 10 at the 80-th and 90-th epoch of total 100 epochs. The weights were initialized with Xavier initialization \cite{glorot2010understanding} and the mini-batch size was set to 8. The weighting parameters $\lambda$ for the total temporal loss in Eq. \eqref{eq8} were empirically set to $\lambda_{R}=1$, $\lambda_{TM}^{1}=1$, $\lambda_{TMM}=1$, $\lambda_{TD}=0.1$, $\lambda^{2}=1$ and $\lambda_{TM}^{2}=0.1$. The weighting parameters for the multi-scale loss in $L=\sum_{l=1}^{3}\lambda_{l}\cdot L_{T}^{l}$ had to be set carefully, since with certain combinations such as $\lambda_{1}=1/3$, $\lambda_{2}=1/3$ and $\lambda_{3}=1/3$, a performance drop was observed compared to a single-scale architecture. We found that more emphasis must be imposed on the lower levels, which is consistent with previous work, and empirically, the best combination was $\lambda_{1}=4$, $\lambda_{2}=2$ and $\lambda_{3}=1$. This is because having an accurate reconstruction to start with is important for the later levels.

\begin{table}
\begin{center}
\caption{Ablation study on the temporal loss}
\label{table:1}
\scalebox{0.82}{
\begin{tabular}{ c|c|c|c|c|c|c }
\hline\hline
 & (a) & (b) & (c) & (d) & (e) & (f) \\
\hline\hline
$L_{R}^{1}$ & \cmark & \cmark & \cmark & \cmark & \cmark & \cmark \\
\hline
$L_{TM}^{1}$ & \xmark & \cmark & \cmark & \cmark & \cmark & \cmark \\
\hline
$L_{TMM}$ & \xmark & \xmark & \cmark & \cmark & \cmark & \cmark \\
\hline
$L_{TD}$ & \xmark & \xmark & \xmark & \cmark & \cmark & \cmark \\
\hline
$L_{R}^{2}, L_{TD}^{2}$ & \xmark & \xmark & \xmark & \xmark & \cmark & \cmark \\
\hline
$L_{TM}^{2}$ & \xmark & \xmark & \xmark & \xmark & \xmark & \cmark \\
\hline\hline
VS-\textit{P} & 36.50 & 36.44 & 36.52 & 36.48 & 36.96 & \textbf{36.99} \\
\hline
S-\textit{P} & \textbf{49.49} & 49.25 & 48.94 & 48.91 & 49.29 & 49.18 \\
\hline\hline
VS-\textit{S} & 0.9632 & 0.9626 & 0.9635 & 0.9630 & 0.9661 & \textbf{0.9662} \\
\hline
S-\textit{S} & \textbf{0.9932} & \textbf{0.9932} & 0.9924 & 0.9925 & 0.9929 & 0.9927 \\
\hline\hline
\multicolumn{7}{l}{*VS: VFI-SR, S: SR, \textit{P}: PSNR (dB), \textit{S}: SSIM}
\end{tabular}}
\end{center}
\end{table}

\begin{table}
\begin{center}
\caption{Ablation study on the temporal loss in a single scale architecture}
\label{table:4-1}
\scalebox{0.88}{
\begin{tabular}{ c|c|c|c|c|c|c }
\hline\hline
 & (a) & (b) & (c) & (d) & (e) & (f) \\
\hline\hline
$L_{R}^{1}$ & \cmark & \cmark & \cmark & \cmark & \cmark & \cmark \\
\hline
$L_{TM}^{1}$ & \xmark & \cmark & \cmark & \cmark & \cmark & \cmark \\
\hline
$L_{TMM}$ & \xmark & \xmark & \cmark & \cmark & \cmark & \cmark \\
\hline
$L_{TD}$ & \xmark & \xmark & \xmark & \cmark & \cmark & \cmark \\
\hline
$L_{R}^{2}, L_{TD}^{2}$ & \xmark & \xmark & \xmark & \xmark & \cmark & \cmark \\
\hline
$L_{TM}^{2}$ & \xmark & \xmark & \xmark & \xmark & \xmark & \cmark \\
\hline\hline
VS-\textit{P} & 36.34 & 36.28 & 36.36 & 36.19 & 36.72 & \textbf{36.78} \\
\hline
S-\textit{P} & \textbf{49.93} & 49.71 & 49.23 & 49.07 & 49.45 & 49.47 \\
\hline\hline
\multicolumn{7}{l}{*VS: VFI-SR, S: SR, \textit{P}: PSNR (dB)}
\end{tabular}}
\end{center}
\end{table}

\subsection{Ablation Study}
\subsubsection{Temporal Loss}
We conducted an ablation study on the components of the temporal loss to analyze their effect. Table \ref{table:1} shows the average PSNR/SSIM performance of the predicted VFI-SR frames and the SR frames for in-depth analysis. This experiment was performed with the multi-scale architecture \textit{without} using the optical flow $f_{t}$ and the warped frames $g_{t}$ inputs, to solely examine the effect of the temporal loss without any additional motion cues. Additionally, we conducted another ablation study on the temporal loss with a \textit{single scale} (not multi-scale) U-Net architecture in Table \ref{table:4-1} to investigate the effect of the temporal loss in simpler CNN architectures. In this experiment as well, $f_{t}$ and $g_{t}$ inputs were not used.

Firstly, the overall PSNR/SSIM values for the SR frames are higher than those of the VFI-SR frames in both Table \ref{table:1} and Table \ref{table:4-1}, because VFI-SR is a more complex joint task where spatio-temporal up-scaling must be performed simultaneously, whereas for SR, only the spatial resolution is up-scaled. Secondly, the usage of the losses $L_{R}^{2}$ and $L_{TD}^{2}$ related to the sample $\{x_{-2\Delta t}, x_{0}, x_{+2\Delta t}\}$ with temporal stride 2 in column (e) of both tables, forces the FISRnet to produce improved reconstruction accuracy by effectively regularizing the temporal relations with 0.48 dB and 0.53 dB gain over column (d) in Table \ref{table:1} and Table \ref{table:4-1}, respectively. Considerable performance gains in SR can be also observed, with 0.38 dB gain from column (d) to (e) in both tables. With $L_{TM}^{2}$ (the temporal matching loss with temporal stride 2 frames) additionally included as in column (f) of Table \ref{table:1} and Table \ref{table:4-1}, 0.51 dB and 0.59 dB gain in PSNR is obtained for joint VFI-SR, respectively, by comparing to column (d). 

Although the final temporal loss improves the prediction accuracy of the VFI-SR frames by enforcing temporal regularization, there exists a performance trade-off between the predicted VFI-SR and SR frames in both cases. The temporal loss adds regularization in the temporal sense at the cost of lowered accuracy for SR predictions. However, we focus on enhancing the joint VFI-SR performance to increase the overall temporal coherence of the final video results. Fig. \ref{fig:5} shows the visual comparison of the VFI-SR frames with and without the temporal loss in both architectures. It is clear that incorporating the temporal loss helps to enhance the edge details and structural construction of objects in both cases with and without the multi-scale structure.

\begin{table}
\begin{center}
\caption{Ablation study on the network architecture. Each component is accumulated from the top to the bottom.}
\label{table:2}
\scalebox{0.9}{
\begin{tabular}{ c|c|c|c|c }
\hline\hline
Network Variants & VS-\textit{P} & S-\textit{P} & VS-\textit{S} & S-\textit{S} \\
\hline\hline
Baseline & 36.34 & \textbf{49.93} & 0.9629 & \textbf{0.9935} \\
\hline
+Temporal Loss & 36.78 & 49.27 & 0.9656 & 0.9931 \\
\hline
+Multi-scale & 36.99 & 49.18 & 0.9662 & 0.9927 \\
\hline
+Optical flow $f_{t}$ & 37.05 & 48.90 & 0.9635 & 0.9922 \\
\hline\hline
+Warped images & \multirow{2}{*}{\textbf{37.66}} & \multirow{2}{*}{47.74} & \multirow{2}{*}{\textbf{0.9740}} & \multirow{2}{*}{0.9918} \\
$g_{t}$ (FISRnet) &&&&\\
\hline\hline
\multicolumn{5}{l}{*VS: VFI-SR, S: SR, \textit{P}: PSNR (dB), \textit{S}: SSIM}
\end{tabular}}
\end{center}
\end{table}

\begin{figure}
\centering
\includegraphics[width=\columnwidth]{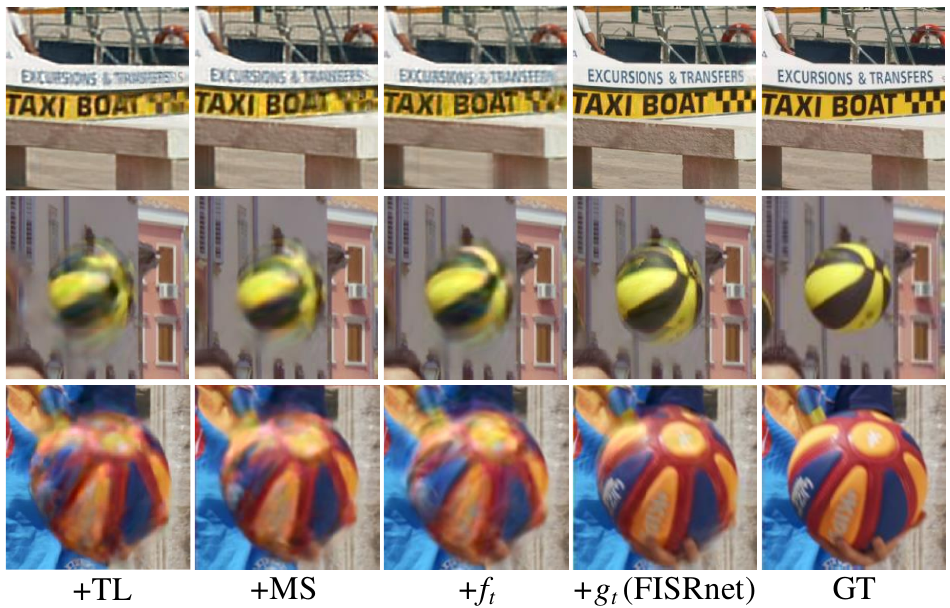}
\caption{Effect of the Network Architecture Components}
\label{fig:6}
\end{figure}

\begin{figure*}
\centering
\includegraphics[scale=0.62]{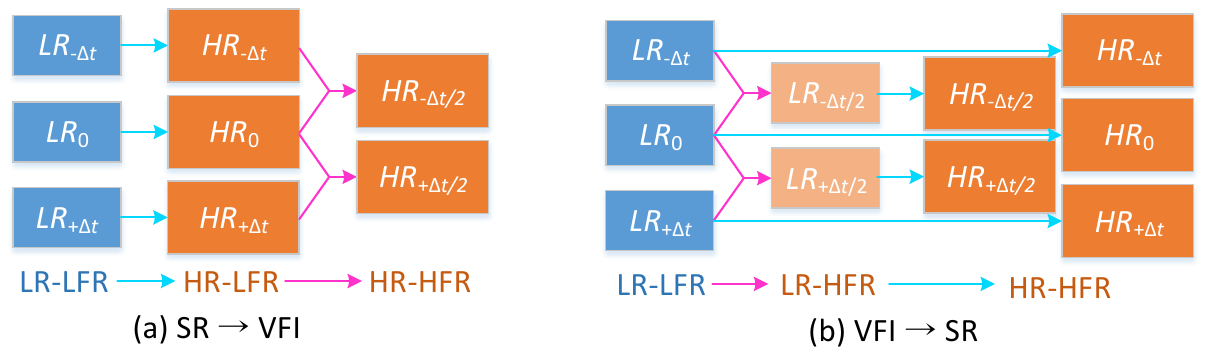}
\caption{Two types of cascaded structures for performing joint VFI-SR}
\label{fig:7}
\end{figure*}

\begin{table*}
\begin{center}
\caption{Quantitative Comparison}
\label{table:3}
\scalebox{0.9}{
\begin{tabular}{ c|c|c|c|c|c|c }
\hline\hline
Order & \multicolumn{2}{c|}{SR $\rightarrow$ VFI} & \multicolumn{2}{c|}{VFI $\rightarrow$ SR} & \multicolumn{2}{c}{Joint}\\
\hline
VFI Method & CyclicGen & CyclicGen & CyclicGen & CyclicGen & \multirow{2}{*}{FISR-Baseline} & \multirow{2}{*}{\textbf{FISRnet (Ours)}}\\ 
\cline{1-5}
SR Method & EDSR & LapSRN & EDSR & LapSRN &&\\ 
\hline\hline
VFI-SR PSNR (dB)&36.15&36.13&36.24&36.23&36.34&\textbf{37.66}\\
\hline
SR PSNR (dB)&49.01&48.88&49.01&48.88&\textbf{49.93}&47.74\\
\hline
Total PSNR (dB)&41.66&41.60&41.71&41.65&\textbf{42.16}&42.00\\
\hline\hline
\end{tabular}}
\end{center}
\end{table*}

\subsubsection{Network Architecture}
Another ablation study was conducted on the architecture components as shown in Table \ref{table:2}. We set the baseline network by excluding the multi-scale feature, optical flows ($f_{t}$) and warped frames ($g_{t}$) from the final FISRnet trained \textit{without} the temporal loss. Each component is accumulatively added from the top row to the bottom row, starting from the baseline network. The temporal loss is again effective, showing 0.44 dB performance gain in PSNR. The multi-scale component also helps to boost the performance, since large motions can be effectively handled in the lower levels of the U-Net with larger receptive fields, guiding the upper level network to learn more efficiently from the coarsely predicted results in this structure. Although the optical flow information results in a marginal performance gain of 0.06 dB in PSNR, additionally providing the warped images as motion information is highly beneficial for VFI-SR, yielding 0.67 dB gain in PSNR if both the optical flow and the warped images are stacked with the input frames in the multi-scale architecture with the temporal loss. Moreover, the components of FISRnet boosts the qualitative performance of the VFI-SR frames as shown in Fig. \ref{fig:6}. The final FISRnet with all components is able to restore the small letters on the boat, and catch the shapes and patterns of the balls and fingers in Fig. \ref{fig:6}. 

\subsection{Comparison with Other Methods}
Since there are no existing joint VFI-SR methods, we conduct an experiment with the cascade of existing VFI and SISR methods with our 4K 60 fps test set. There can be two variations of the cascade connections as shown in Fig. \ref{fig:7}. In the first variation as shown in Fig. \ref{fig:7} (a), SR can be performed first to enlarge the spatial resolution of the LR frames, resulting in HR-LFR frames, then VFI can be performed on the up-scaled frames to obtain the HR middle frames for finally generating the HR-HFR video outputs. As for the second variation shown in Fig. \ref{fig:7} (b), the LR middle frames can be produced first to increase the temporal resolution, resulting in LR-HFR frames, and then SR can be performed on all LR frames to generate the final HR-HFR video outputs. For the compared methods, we select the recent CyclicGen \cite{liu2019deep} for the VFI method, and cascade EDSR \cite{lim2017enhanced} or LapSRN \cite{LapSRN} as the SR method. For all methods, we used the official codes provided by the authors. 

\subsubsection{Quantitative Comparison}
The quantitative comparison for the FISRnet and the cascaded methods are given in Table \ref{table:3}. For the cascade orders, performing VFI followed by SR (VFI $\rightarrow$ SR) seems to generally show better performance, since in the perspective of the VFI method, it is easier to capture the motion along the temporal evolution of the 2K LR frames (VFI $\rightarrow$ SR) than along the up-scaled 4K HR frames (SR $\rightarrow$ VFI), where the absolute motion displacement is larger. Our proposed FISRnet outperforms the four cascaded combinations for the VFI-SR frames with at least 1.42 dB gain in terms of PSNR. Due to the trade-off between VFI-SR and SR performance, the baseline architecture of FISR (FISR-Baseline) shows better performance for SR, outperforming EDSR by 0.92 dB.

\subsubsection{Qualitative Comparison}
\begin{figure*} [t]
\centering
\includegraphics[width=\textwidth]{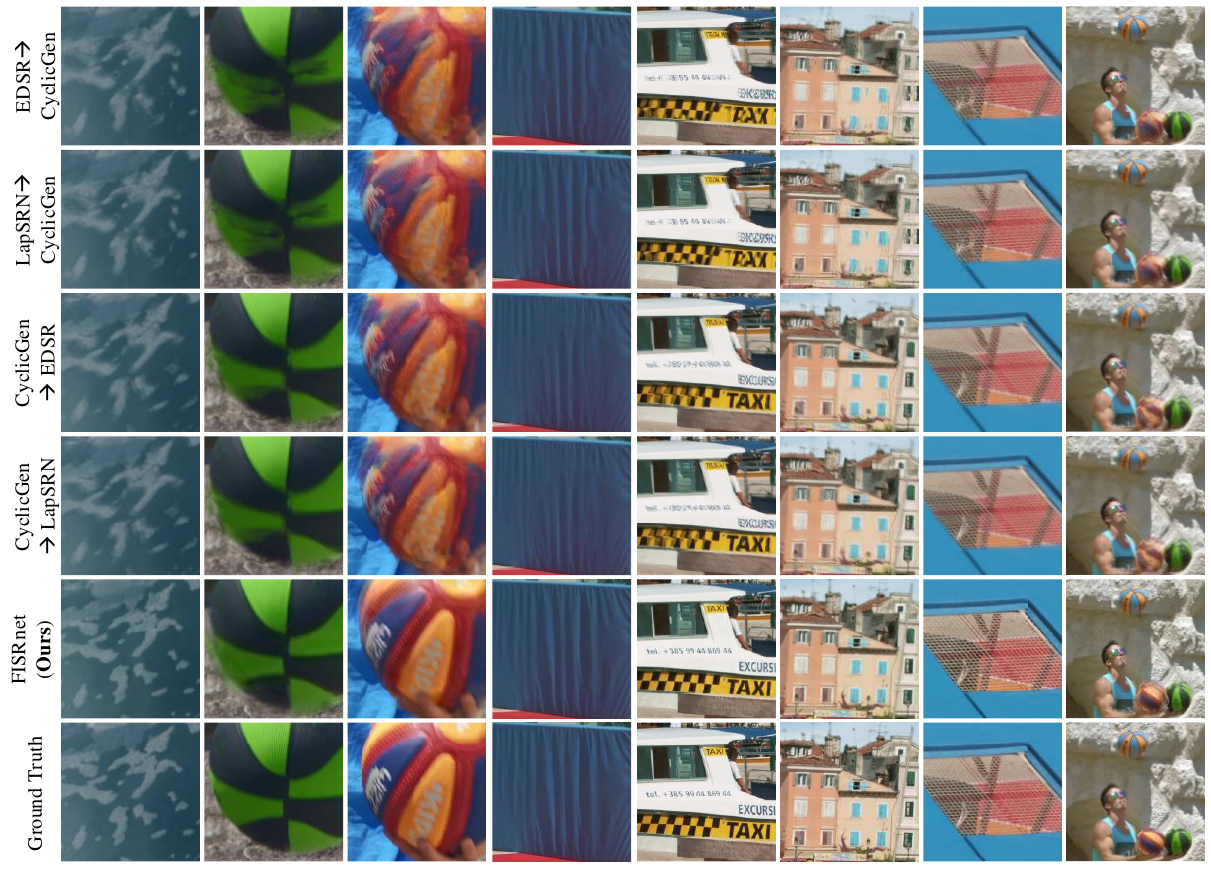}
\caption{Qualitative Comparison}
\label{fig:8}
\end{figure*}

The qualitative comparison of the VFI-SR frames is given in Fig. \ref{fig:1} and Fig. \ref{fig:8}. FISRnet accurately reconstructs the objects with realistic textures and sharp edges. Our method is able to capture the texture of the water waves and reconstruct small letters on the ball and the boat in Fig. \ref{fig:1} and Fig. \ref{fig:8}. Furthermore, performing VFI followed by SR generates better structural context ($2^{\text{nd}}$ and $5^{\text{th}}$ column in Fig. \ref{fig:8}) but often produces blurry edges, while SR followed by VFI restores sharper edge details ($4^{\text{th}}$ and $6^{\text{th}}$ column in Fig. \ref{fig:8}) at the cost of less accurate structural reconstructions. In the latter case, the motion displacement seems to have exceeded the maximum motion that the network \cite{liu2019deep} can handle, due to the large resolution (4K) inputs.

\subsection{Runtime Evaluation}
The testing runtime of FISRnet is average 2.73 seconds, for one input test sample of three 2K (1920$\times$1080) resolution frames that generates two 4K (3840$\times$2160) VFI-SR frames and one 4K SR frame at once, with an NVIDIA TITAN Xp GPU.

\section{Conclusion}
In this paper, we first defined a novel problem of joint VFI-SR to directly synthesize high quality HR HFR frames from LR LFR input frames, which can be applied for the direct conversion of 2K 30 fps videos to 4K 60 fps videos. This is a very useful means to generate high quality visual content for premium displays. However, joint VFI-SR is a difficult task, where the spatio-temporal up-scaling must be performed simultaneously to produce non-existent up-scaled frames. We proposed a three-level multi-scale U-Net-based network, called FISRnet, to handle the large motion present in the high resolution data of 2K resolution inputs, trained via the proposed temporal loss with the multiple data sample training strategy that allows for a more stable temporal regularization. Applying the temporal loss exploits the temporal relations existing across the multiple data samples, helping the FISRnet to sharpen the edges and construct the correct shapes of diverse objects. Besides, the temporal loss and the multiple data sample training can be applied to any video-related vision task. We analyzed the effect of the temporal loss and the components of the network architecture with various ablation studies in the Experiment Section, and also demonstrated that our FISRnet outperforms the cascades of existing state-of-the-art VFI and SISR methods. The official Tensorflow code is available at \textit{https://github.com/JihyongOh/FISR}.

\section{Acknowledgement}
This work was supported by Institute for Information \& communications Technology Promotion (IITP) grant funded by the Korea government (MSIT) (No. 2017-0-00419, Intelligent High Realistic Visual Processing for Smart Broadcasting Media).

\bibliography{AAAI-KimS.4711.bbl}
\bibliographystyle{aaai}

\end{document}